# On Detecting Noun-Adjective Agreement Errors in Bulgarian Language Using GATE

**Nadezhda Borisova, Grigor Iliev, Elena Karashtranova**

*South-West University, Blagoevgrad, Bulgaria*

***Abstract***:  In this article, we describe an approach for automatic detection of noun-adjective agreement errors in Bulgarian texts by explaining the necessary steps required to develop a simple Java-based language processing application. For this purpose, we use the GATE language processing framework [9], which is capable of analyzing texts in Bulgarian language and can be embedded in software applications, accessed through a set of Java APIs. In our example application we also demonstrate how to use the functionality of GATE to perform regular expressions over annotations for detecting agreement errors in simple noun phrases formed by two words – attributive adjective and a noun, where the attributive adjective precedes the noun. The provided code samples can also be used as a starting point for implementing natural language processing functionalities in software applications related to language processing tasks like detection, annotation and retrieval of word groups meeting a specific set of criteria.

## 1. INTRODUCTION

The adjective in Bulgarian language must agree with the noun in number, gender, and definiteness. For example, in the noun phrase "щастливо деца" ("a happy children") the adjective and the noun do not agree in number – the adjective "щастливо" is singular and the noun "деца" is plural. In the noun phrase "щастлив дете" ("a happy child") the adjective and the noun do not agree in gender – the adjective "щастлив" is masculine and the noun "дете" is neuter. While these types of errors are rarely made by native speakers of Bulgarian, for people learning Bulgarian as a foreign language, this is not the case. This is why tools for automatic detection of such errors can be very useful.

Our choice to use GATE for this purpose is based mainly on the fact that it is a mature open source project with over 15 years of development in the field of human language processing [6] and provides extensive support

Mathematics and Informaticsfor integration and interoperability with most of the other key systems and tools in this field [2].

Taking into account the specifics of the GATE framework, the task of automatic detection of noun-adjective agreement errors can be divided into a set of related subtasks:

- Breaking a text into sentences.
- Breaking a text into tokens (words, numbers, etc.).
- Tagging individual tokens with their respective part of speech (POS) tags.
- Tagging adjective–noun pairs that do not agree in gender, number and definiteness.

The first three subtasks are standard natural language processing (NLP) tasks – sentence splitting, tokenization and POS tagging [8]. The last subtask we will handle using the JAPE (Java Annotation Patterns Engine) transducer, which is part of the GATE system and provides finite state transduction over annotations based on regular expressions [1]. For more information about JAPE see [5].

Note that in this article we focus on the use of the GATE APIs for performing specific language processing tasks, which requires a basic understanding of the Java programming language. In addition, a basic understanding of regular expressions and writing JAPE rules is also required.

## 2. EXTERNAL LIBRARY DEPENDENCIES

To make the GATE APIs available to a Java based application, one need to deploy the following Java extensions:

- `$GATE_HOME/bin/gate.jar`
- all JAR files in `$GATE_HOME/lib`

Where `$GATE_HOME` is the absolute path of the GATE's root directory [1].

It is also mandatory to initialize the GATE library before first use. This is done by calling the static method `gate.Gate.init` [7].

## 3. LOADING CREOLE PLUGINS

Before using a certain processing resource in GATE, first the CREOLE plugin containing the resource should be loaded. In our example we will use four processing resources – LingPipe Tokenizer PR, LingPipe Sentence Splitter PR, LingPipe POS Tagger PR and JAPE-Plus Transducer. The first three resources are part of the LingPipe plugin. The last one is part of the

181

JAPE_Plus plugin. Our choice to use the LingPipe plugin is based on the fact that currently this is the only plugin, included in the GATE distribution, with POS tagger which includes models for Bulgarian language. We must, however, note that there is no restriction in using tokenizers and sentence splitters from another plugins in conjunction with the POS tagger from the LingPipe plugin.

To load the LingPipe and JAPE_Plus plugins, we need to register the corresponding CREOLE directories. This is done by issuing the following code [1]:

```
java.io.File f = new java.io.File (
    gate.Gate.getPluginsHome(), "LingPipe"
);
gate.Gate.getCreoleRegister().registerDirectories (
    f.toURI().toURL()
);

f = new java.io.File (
    gate.Gate.getPluginsHome(), "JAPE_Plus"
);
gate.Gate.getCreoleRegister().registerDirectories (
    f.toURI().toURL()
);
```

## 4. LOADING PROCESSING RESOURCES

The next step, after loading the plugins containing the aforementioned resources which will be used in our example, is to load the actual processing resources.

As pointed out in the introduction, we need processing resources for the following tasks:
- Breaking a text into sentences.
- Breaking a text into tokens.
- Tagging the individual tokens with their respective part of speech tags.
- Tagging adjective–noun pairs that do not agree in gender, number and definiteness.

The Sentence Splitter processing resource is used for breaking the text into sentences. To create it with default parameters, we use the following code:

```
gate.FeatureMap params =
    gate.Factory.newFeatureMap();

gate.ProcessingResource splitter =
```

```
      (gate.ProcessingResource)gate.Factory.createResource (
         "gate.lingpipe.SentenceSplitterPR", params
      );
```

To create a Tokenizer processing resource with default parameters, we use the following code:

```
gate.FeatureMap params = gate.Factory.newFeatureMap();
gate.ProcessingResource tokenizer =
   (gate.ProcessingResource)gate.Factory.createResource (
      "gate.lingpipe.TokenizerPR" , params
   );
```

Once the input text is broken into tokens, the LingPipe POS Tagger processing resource can be used for tagging the individual tokens with their respective part of speech tags. For this purpose, we need to provide a model for Bulgarian language as an initialization parameter [1, 7]:

```
String model =
    "LingPipe/resources/models/bulgarian-full.model";
java.io.File f =
    new java.io.File(gate.Gate.getPluginsHome() , model);

gate.FeatureMap params = gate.Factory.newFeatureMap();
params.put("modelFileUrl", f.toURI().toURL());

gate.ProcessingResource tagger =
    (gate.ProcessingResource)gate.Factory.createResource (
        "gate.lingpipe.POSTaggerPR" , params
    );
```

Finally, we create a JAPE-Plus Transducer, which will be used to detect the adjective–noun pairs that do not agree in gender, number and definiteness. As an initialization parameter we need to provide the grammar file containing our custom JAPE rules:

```
gate.FeatureMap params = gate.Factory.newFeatureMap();
java.io.File f = new java.io.File("/jape/Example.jape");
params.put("grammarURL", f.toURI().toURL());

String transducerClass =
    gate.jape.functest.TransducerType.PLUS.getFqdnClass();

gate.ProcessingResource japePlus =
    (gate.ProcessingResource)gate.Factory.createResource (
       transducerClass, params
    );
```

We will discuss the content of Example.jape file in Section 6.

## 5. CREATING AND RUNNING A CORPUS PIPELINE

One of the ways to run the already created processing resources in sequence is to use a *corpus pipeline*. For this purpose, we will use the class SerialAnalyserController [1]:

```
gate.creole.SerialAnalyserController controller;
controller = (gate.creole.SerialAnalyserController)
    gate.Factory.createResource (
        "gate.creole.SerialAnalyserController",
        gate.Factory.newFeatureMap(),
        gate.Factory.newFeatureMap()
    );
```

The next step is to add the already created processing resources to the controller:

```
controller.add(splitter);
controller.add(tokenizer);
controller.add(tagger);
controller.add(japePlus);
```

Note that the order in which the processing resources are added is important. For example, the POS tagger requires as an input the 'Token' and 'Sentence' annotations, which are created by the tokenizer and the sentence splitter. The JAPE transducer will require each 'Token' annotation to have a 'category' feature, set with the part of speech tag of the respective token, which is done by the POS Tagger. Note that, for example, by "'Token' annotation", we mean annotation of type 'Token'.

The controller that we use in our example accepts only processing resources that process documents, also called *language analysers*. Hence, the controller will require a *corpus*, which is basically a collection of documents. When executed, the controller runs every analyser over each document in the corpus. For our purposes we will create a corpus with a single document in it [7]:

```
gate.Corpus corpus;
corpus = gate.Factory.newCorpus("Test Corpus");
java.io.File f = new java.io.File("/texts/bg.txt");
gate.Document doc;
doc = gate.Factory.newDocument(f.toURI().toURL());
corpus.add(doc);
```

Mathematics and Informatics

Next, we set the controller to use the already created corpus and execute the added processing resources in sequence.

```
controller.setCorpus(corpus);
controller.execute();
```

## 6. WRITING THE JAPE GRAMMAR RULES

In this section, as an example, we show how to create a JAPE grammar rule, which matches all adjective–noun pairs, where the adjective is in plural form and the noun is in singular form. The matched adjective–noun pairs are then annotated as type 'PSAgrError'. We will call this rule PluralSingularPair.

Note that it is essential to specify at the start of the JAPE grammar the input annotations, against which the rules will be matched. In our case, we only need to iterate through the annotations of type 'Token'. So, the following line should be included in the JAPE grammar file "Example.jape":

```
Input: Token
```

In our example, the parts of speech are annotated by the LingPipe POS tagger using the BulTreeBank tagset (BTB-TS) scheme [4] and are added to the document as 'category' features of the 'Token' annotations. According to the BTB-TS tagset specification, if the first character of a POS tag is 'A', the respective token is an adjective, if it is 'N', the respective token is a noun. For nouns, the fourth character of the POS tag determines the number of the noun – 's' for singular, 'p' for plural. For adjectives, it is the third character. Hence, for example, to match all annotations of type 'Token' with 'category' feature, whose value is a string that starts with the capital letter 'A', followed by an arbitrary character and the lower-case letter 'p', we use the following pattern:

```
{ Token.category =~ "^A.p" }
```

To match all annotations of type 'Token' with feature 'category', whose value is a string that starts with the capital letter 'N', followed by two arbitrary characters and the lower-case letter 's', we use the following pattern:

```
{ Token.category =~ "^N..s" }
```

Note that the regular expression operator "^" matches the beginning of the line and the operator "." matches any single character [3].

Using these two patterns, we create a rule, which matches all adjective–noun pairs, where the adjective is in plural form and the noun is in singular form:

```
Rule: PluralSingularPair
Priority: 20
(
    { Token.category =~ "^A.p" }
    { Token.category =~ "^N..s" }
): pair
-->
:pair.PSAgrError = { rule = "PluralSingularPair" }
```

Note that in the last line of the rule, the temporary label "pair" is renamed to "PSAgrError", thus creating a new annotation of type 'PSAgrError'. Rules for covering other types of agreement errors can be implemented in the same manner.

## 7. CONCLUSION

As shown above, developing a language processing application using GATE does not require a significant learning curve. The time and effort invested is negligible and well-worth the benefits of the derived functionality [2]. The ability to perform regular expressions over annotations using the full power of pattern matching provides a wide range of techniques for solving all kinds of real problems related to computer processing of human language. Additionally, the limitations imposed by the JAPE grammar can be overcome by invoking custom Java code inside the JAPE rules.

We conclude that the GATE framework with its multilingual support and language-independent components is a reasonable choice for developing a software for automatic detection and correction of a wide range of grammatical errors in Bulgarian texts, which will be of great help for people learning Bulgarian as a foreign language.

## 8. ACKNOWLEDGEMENT

This research was supported by the South-West University "Neofit Rilski" grant SRP-A5/12.